\documentclass[sigconf]{acmart}
\usepackage [autostyle]{csquotes}
\usepackage{algorithm}
\usepackage[noend]{algpseudocode}
\AtBeginDocument{%
  \providecommand\BibTeX{{%
    \normalfont B\kern-0.5em{\scshape i\kern-0.25em b}\kern-0.8em\TeX}}}

\copyrightyear{2023}
\acmYear{2023}
\setcopyright{rightsretained}
\acmConference[GECCO '23 Companion]{Genetic and Evolutionary Computation Conference Companion}{July 15--19, 2023}{Lisbon, Portugal}
\acmBooktitle{Genetic and Evolutionary Computation Conference Companion (GECCO '23 Companion), July 15--19, 2023, Lisbon, Portugal}\acmDOI{10.1145/3583133.3590725}
\acmISBN{979-8-4007-0120-7/23/07}

\begin{document}

\title{The Struggle for Existence: Time, Memory and Bloat}
\author{John Stevenson}
\orcid{0000-0001-8518-9997}
\affiliation{%
  \institution{Long Beach Institute}
  \streetaddress{16 Michigan St}
  \city{Long Beach}
  \state{New York}
  \country{USA}
  \postcode{11561}
}
\email{jcs@alumni.caltech.edu}


\begin{abstract}
	Combining a spatiotemporal, multi-agent based model of a foraging ecosystem with linear, genetically programmed rules for the agents' behaviors results in implicit, endogenous, objective functions and selection algorithms based on \textquote{natural selection}. Use of this implicit optimization of genetic programs for study of biological systems is tested by application to an artificial foraging ecosystem, and compared with established biological, ecological, and stochastic gene diffusion models. Limited program memory and execution time constraints emulate real-time and concurrent properties of physical and biological systems, and stress test the optimization algorithms. Relative fitness of the agents' programs and efficiency of the resultant populations as functions of these constraints gauge optimization effectiveness and efficiency. Novel solutions confirm the creativity of the optimization process and provide an unique opportunity to experimentally test the neutral code bloating hypotheses. Use of this implicit, endogenous, evolutionary optimization of spatially interacting, genetically programmed agents is thus shown to be novel, consistent with biological systems, and effective and efficient in discovering fit and novel solutions. 
		
\end{abstract}



\maketitle
\section{Introduction}

Classification of biological, sociological, and ecological models includes minimal and synthetic models of systems \cite{roughgarden}. Synthetic models match the macroscopic results of the model to empirical data or desired results \cite{grimm}, provide explanatory rules \cite{epstein} and, when used with genetic programming, can discover novel rules \cite{vu}. In contrast, minimal models do not attempt to calibrate to an empirical objective function. Rather, a population of agents interacts and freely evolves within an environment. Some models in this category have external objective functions and selection processes \cite{adami1994}. Others, as used here, apply the \textcquote{gause}{struggle for existence} endogenously within the simulation \cite{axtell,ray1992}.

The spatiotemporal, multi-agent based model (ABM) of a foraging ecosystem, with agents competing for the same resources and space, selects agents for reproduction based on their survival, resources, and fertility. When combined with agents running genetic programs to determine their actions, an explosion of novel and complex agent behaviors emerge \cite{stevensonGP}. While there has been recent interest in genetically programming agent behaviors within ABMs for systemic models of systems \cite{vu},
application of this minimal model of a system to address the ecological concerns of movement based community assembly is novel \cite{schlagel,ray1992,adami1994} . The genetic programming objective function and the selection algorithm are implicit and endogenous to the model.
No separate processing for evaluation, selection, or multi-goal optimization is required. 
This optimization of agents behaviors preserves the biological, ecological, and stochastic gene population characteristics \cite{stevensonGP}. By designing these agents with constraints on the number of instructions per action cycle (compute capacity) and program size, real-time performance limits and concurrent actions by large, real-world, interacting populations are simulated. The effects these constraints of execution time and program size have on relative fitness, population efficiency, and bloat are then readily measured and provide new data on limits of real-world agents.\footnote{Population efficiency is inferred from realized carry capacity \cite{murray}. Relative fitness is determined by competitive exclusion \citep{chessMech}. The predicted steady state carry capacity is based on the resources flowing into the landscape \cite{murray,stevenson}}
Specific experiments are then run whose results empirically support the \textquote{neutral code is protective} hypothesis \cite{banzhaf} and suggest potential advantages of 
preadaptation \cite{wilsonPreAdapt}.


\section{Methods}

The underlying spatiotemporal, multi-agent-based model (uABM) is a minimum model of a system \cite{roughgarden} based on Epstein and Axtell's classic Sugarscape \cite{axtell,stevenson}. For these experiments, the only uABM parameters (contained in the agent's genome) that are varied across runs are infertility and birth cost. Birth cost is the parent's required surplus resources that are consumed during reproduction. Given the birth cost and free space constraints are met, the probability of reproduction, $p_{f}$, is expressed as infertility $ f = \frac{1}{p_{f}}$. Reproduction is haploid (cloning) with a stochastic, single-point mutation of an agent's genome. The agents interact on an equal opportunity (flat) landscape of renewing resources. The landscape parameters remain constant during a run. The population dynamics that emerge from this simple underlying model have been shown to agree with time delayed logistic growth models for single species 
\cite{murray,stevenson}
, stochastic gene diffusion models \cite{ewens,stevenson}, and modern coexistence theory \citep{chessMech,stevensonX}. Agents only die when their metabolism exhausts their current resources, otherwise they are immortal.

The uABM provides the structure for the genetic programming of the agents' behaviors. A simple linear genetic programming language replicating the uABM agent rules was designed and integrated into a genetically-programmed ABM (gpABM). The genome with the agents' characteristics and the landscape parameters remain constant during runs for the gpABM. Each agent's program is contained in a 32 character string which contains the registers and instructions executed on each agent's action cycle. Five characters are used for registers leaving up to 27 characters for the program.  These instructions and registers are described in Table \ref{table:apl}. The action cycle for this gpABM is depicted in Algorithm \ref{action}.

\begin{table}[h!]
	\caption{Agent Registers and Instruction Set}
	\begin{center}
		\resizebox{\columnwidth}{!}{
			\begin{tabular}{|c|c|c|c|c|}
				\hline
				Name & Address & Function & Values & Description \\
				\hline
				nextI & 1-2 & register & 05-31 & address of next instruction  \\
				bDir & 3 & register & UDLRZ &  best seen direction (Z$=$no data)\\
				bDis & 4 & register & 0-9 & best seen distance \\
				bRes & 5 & register & 0-9 & best seen resources \\
				inst & 6-32 & program & UDLRMX & executeable instruction\\
				\hline
			\end{tabular}
		}
		\bigbreak
		\resizebox{\columnwidth}{!}{
			\begin{tabular}{|c|c|c|c|}
				\hline
				Instr & Description & Action/Test & Result\\
				\hline
				U & look up & find cell max resource above $>$ bRes & store in bDir,bDis,bRes\\
				D & look down & find cell max resource below $>$ bRes & store in bDir,bDis,bRes\\
				L & look left & find cell max resource left $>$ bRes & store in bDir,bDis,bRes\\
				R & look right & find cell max resource right $>$ bRes & store in bDir,bDis,bRes\\
				M & move & fetch bDis, bDir, if 'Z' random values  & move bDis,bDir \\
				X & reproduce & space, birth costs allow reproduction & place new agent in cell \\
				\hline
			\end{tabular}
		}
		\label{table:apl}
	\end{center}
\end{table}

The number of instructions that can be executed per each agent's action cycle, called compute capacity, is part of the parameter space that is surveyed. Foraging gains, metabolic costs, and death by starvation occur during the move instruction (Algorithm \ref{action}). 
The results, with initial hand-crafted programs replicating the uABM, have indistinguishable population and genetic dynamics, and agent metrics from the uABM. These replication programs often emerge spontaneously as good solutions and, surprisingly, are sometimes competitively excluded.

\begin{algorithm}
	\begin{algorithmic}[1]
		\small
		\Procedure{action cycle}{}
		\While{action list not empty}
		\State select random agent $a_{i}$, set counter $x=0$
		\While{$x<$compute capacity and surplus $S_{i}>=0$}
		\State execute next instruction, increment $x$
		\EndWhile
		\If{no metabolism expended}
		\State update $S_{i}$ by foraged - metabolism
		\EndIf
		\If{ $S_{i}<0$}
		\State dead ,remove $a_{i}$ from simulation
		\EndIf
		\State remove $a_{i}$ from action list
		\EndWhile
		\EndProcedure
		\Procedure{look}{}
		\State find closest best resource $C_{r}$ in direction $bDir$
		\If{$C_{r} >$ stored best resource $bRes$}
		\State update $bRes,bDir$, and distance 
		\EndIf
		\EndProcedure
		\Procedure{move}{$a_{i},bDist,bDir$}
		\If{target cell unoccupied}
		\State move $bDist$ in $bDir$, update $S_{i}$ by forage - metabolize
		\EndIf
		\EndProcedure
		\Procedure{reproduce}{$a_{i}$}
		\If{1 in $f$, age$>=$puberty, room, and $S_{i}>=$birth cost}
		\State place offspring in empty cell, reduce $S_{i}$ by birth cost
		\If{puberty is zero}
		\State place offspring on action list
		\EndIf
		\EndIf
		\EndProcedure
	\end{algorithmic}
	\caption{\small Action cycle for gpABM}
	\label{action}
\end{algorithm}


While genetic algorithms and genetic programming have a large body of techniques for various objectives when shaping the evolving populations of solutions \cite{koza}, this research uses the implicit optimization of the \textcquote{gause}{struggle to exist} to select agents' programs endogenously within the simulation. When an agent reproduces, a single point mutation will occur in the daughter agent at a constant probability $\frac{1}{\mu}$ per reproduction. If a mutation occurs, a location in the program and a type of mutation are chosen randomly. Three mutation types are implemented: flip to a different random instruction; add a new random instruction if memory space allows; or delete an instruction (if the program is longer than one instruction). There is no cross-over mutation.

\begin{figure}
	\begin{center}
		\resizebox{\columnwidth}{!}{
			\includegraphics[angle=-90,scale=1]{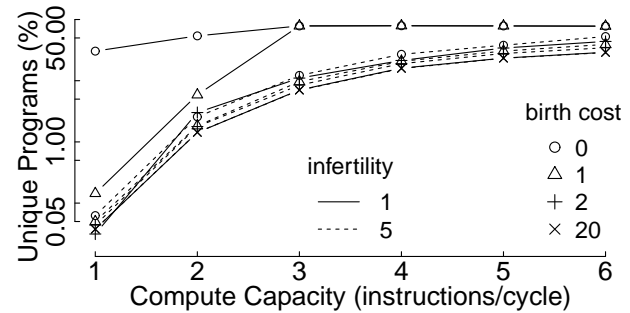}
		}
	\end{center}	
	\caption {\small The fraction of random programs that survive and reproduce through the initial population minimum.}
	\label{fig:phenoStats}
\end{figure}

All runs are initiated with a population of 400 programs with random instructions of random length. Different seeds generate different initial populations and resultant population trajectories. Some infertility and birth cost configurations are so challenging that only a few of the initial random sets of programs were able to generate viable, reproducing populations. Figure \ref{fig:phenoStats} gives the fraction of the initial random population that survives the initial culling and is fertile. Simulation runs are generally stopped at either 10,000 or 50,000 generations, orders of magnitude past the attainment of steady population levels. On these time scales program evolution continues to occur so the stopping point is somewhat arbitrary. 

\section{Results}

Under compute capacity and memory constraints, the performance of this optimization is represented by population efficiency in terms of realized carry capacity \cite{murray,stevenson} in Section 3.1, by relative fitness of individual programs within a population \citep{chessMech,stevensonX} in Section 3.2 (including a novel emergence of a growing infertile caste), and the occurrence or absence of bloat \cite{banzhaf,vu} in Section 3.3 (which uses a novel and surprising set of programs). The ability to grow populations from initial random populations in challenging configurations was shown in Figure \ref{fig:phenoStats}. There is also the efficiency gained by not having separate objective functions and selection processes.

\subsection{Compute Capacity Population Fitness} 
\begin{figure}
	\begin{center}
	\resizebox{\columnwidth}{!}{
		\includegraphics[angle=-90,scale=1.0]{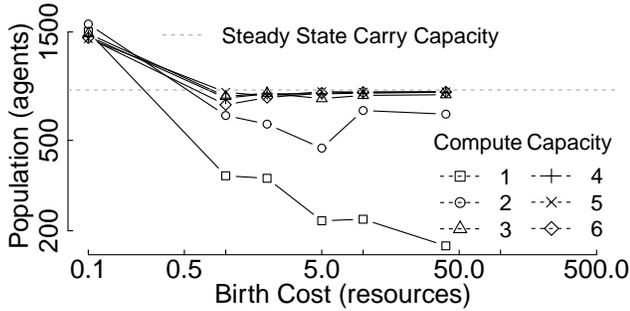}
}		
	\end{center}	
	\caption {\small The realized carry capacity versus birth cost across compute capacities.}
	\label{fig:cc}
\end{figure}

Emergent agent behaviors were significantly affected by the compute capacity of the agents. 
Stable populations at varying population levels emerged for all compute capacities investigated, from one to six. The efficiency of a population has a non-linear relationship with compute capacity and birth cost as shown in Figure \ref{fig:cc}. As birth cost increases, populations with more limited compute capacity become increasingly less efficient. As both Figures \ref{fig:phenoStats} and \ref{fig:cc} show, reducing compute capacity makes life much more difficult for finding solutions, and for the overall efficiency of the population. The unexpected stable population levels well above the steady state carry capacity for birth cost 0 and non-stochastic infertility 1 are discussed in Section 3.3.

\subsection{Memory Size Dependent Fitness}
\begin{figure}
	\begin{center}
		\resizebox{\columnwidth}{!}{
			\includegraphics[angle=-90,scale=0.6]{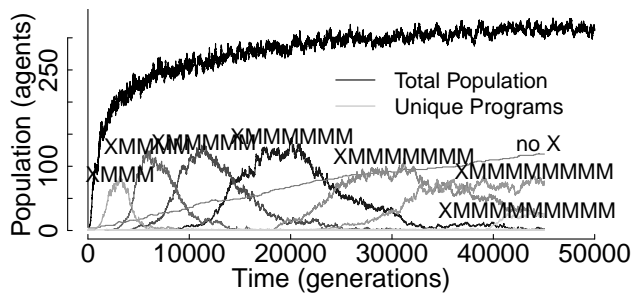}
		}
	\end{center}
	\small	
	\caption {\small Population levels and efficiency slowly increasing as unique mutations drive toward the memory limit.}
	\label{fig:length}
\end{figure}

For configurations with compute capacity 1 and birth costs greater than zero, a simple strategy dominates the remaining configurations of infertility and birth cost. This strategy is of the form 'XMM' with the number of move instructions expanding to eventually fill the agent's allocated program memory. Figure \ref{fig:length} shows both the increase in population level over time as an indication of increasing colony fitness and the invasions of more fit mutations containing additional move instructions. Over time, the optimization will discover more fit, longer programs until the memory capacity is reached. However, at the same time, mutations that delete the reproduction instruction ('X') leave programs (labeled 'no X' in Figure \ref{fig:length}) that cannot reproduce but are slightly more fit than the fertile programs. This unexpected sterile caste is suggestive of eusociality. 

As compute capacity increases, there is time for more efficient search patterns involving look instructions. Cooperative search patterns between two or more distinct programs also emerge \cite{stevensonGP}. These cooperative programs are often more fit and exclude the strategies used in the uABM.
\subsection{Neutral Code Bloat}
Bloat is considered undesirable for its impact on execution times of the simulation, overfitting, and explainability \cite{vu}; though it also may have preadaptation advantages \cite{wilsonPreAdapt}. The generation of bloat may be due to the \textquote{neutral code is protective} hypothesis \cite{banzhaf}. Since the compute capacity is limited in these experiments, bloat carries a fitness penalty for programs that execute it and is, therefore, competitively excluded. 

A novel strategy, however, emerges for  birth cost 0 and non-stochastic infertility of 1 where bloat does occur and stable population levels are well above the steady state carry capacity (Figure \ref{fig:cc}). The calculation of steady state carry capacity is based on the number of agents that can survive on the amount of resources that flow into the landscape each time step \cite{murray}. For these configurations, however, all of these successful strategies die within the first action cycle (once the very brief growth to the maximum population level has occurred). Since none of the agents survive past a single generation, the steady state carry capacity does not apply.


\begin{figure}
	\begin{center}
		\resizebox{\columnwidth}{!}{
			\includegraphics[angle=-90,scale=1.0]{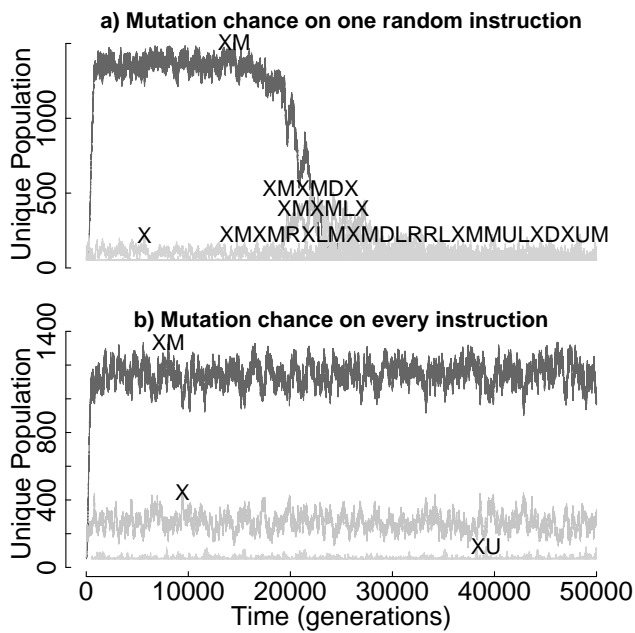}
		}
	\end{center}
	\caption{\small a) Bloat improves fitness when possible mutation is only once per program. b) Bloat provides no advantage when possible mutation is per instruction.}
	\label{fig:bloat}
\end{figure}

This surprising survival strategy allows a test of bloating in a linear GP optimization. The instructions that follow the first compute cycle are never executed and represent neutral code bloating. A representative run (with compute cycle 2 and $\mu = 30$) is shown in Figure \ref{fig:bloat}a. At first, the program 'XM' dominates the population as the most fit result emerging from the initial random programs. After 20,000 generations, however, a phalanx of increasingly longer mutations emerge which quickly exclude the resident small program. The bloated, longer programs are thus demonstrated to be relatively more fit \cite{chessMech} though their executed code is identical. The population level (and therefore its efficiency) remains the same. To confirm whether this improved fitness is due to the \textquote{protection} from mutation of the key instructions that the longer program offers, the mutation algorithm was modified. Figure \ref{fig:bloat}b shows the same run when the probability of mutation is applied to each instruction separately rather than just once for each program. The likelihood that instructions executed in the first action cycle will mutate does not change with program length, thereby eliminating the protection long neutral code programs provide and supporting the \textquote{neutral code is protective} hypothesis \cite{banzhaf}. 

These more fit bloated programs also support the potential advantage of
\emph{de novo gene birth} preadaptation \cite{wilsonPreAdapt}. These long strands of unexecuted code are thus available for complex adaptations under further mutation.

\section{Conclusion}
A spatiotemporal, multi-agent based model of a foraging ecosystem implicitly provides a specific objective function and selection algorithm for genetically programmed agent strategies. 
This implicit, endogenous, evolutionary optimization is consistent with biological systems and shown to be effective in both exploring the solution space and discovering fit, efficient, and novel solutions. Limiting the number of instructions that can be executed during an agent's action cycle emulates real-time and concurrent properties of physical and biological systems, and has significant effects on relative fitness of the individual programs and of the aggregate populations. 
Compute-capacity limits naturally exert selection pressure against bloating. The novel solutions that emerged for zero birth cost and non-stochastic infertility allowed experiments that supported both the \textquote{neutral code is protective} hypotheses and suggest potential advantages of bloating for preadaptation.





\bibliographystyle{ACM-Reference-Format}
\bibliography{feb03_23.bib}


\end{document}